\pdfoutput=1

\documentclass[11pt]{article}

\usepackage{EACL2023}

\usepackage{graphicx}
\usepackage{times}
\usepackage{latexsym}
\usepackage{amsmath}
\usepackage{bbm}
\usepackage{amssymb}
\usepackage{multirow}

\usepackage[T1]{fontenc}

\usepackage[utf8]{inputenc}

\usepackage{inconsolata}
\usepackage{microtype}
\usepackage{makecell}

 \newcommand{\argmax}{\mathop{\rm arg~max}\limits}
 
\title{Semantic Specialization for Knowledge-based Word Sense Disambiguation}

\author{Sakae Mizuki \and Naoaki Okazaki \\
  Tokyo Institute of Technology \\
  \texttt{\{sakae.mizuki@nlp., okazaki@\}c.titech.ac.jp} \\}

\begin{document}
\maketitle
\begin{abstract}

A promising approach for knowledge-based Word Sense Disambiguation (WSD) is to select the sense whose contextualized embeddings computed for its definition sentence are closest to those computed for a target word in a given sentence. This approach relies on the similarity of the \textit{sense} and \textit{context} embeddings computed by a pre-trained language model. We propose a semantic specialization for WSD where contextualized embeddings are adapted to the WSD task using solely lexical knowledge. The key idea is, for a given sense, to bring semantically related senses and contexts closer and send different/unrelated senses farther away. We realize this idea as the joint optimization of the Attract-Repel objective for sense pairs and the self-training objective for context-sense pairs while controlling deviations from the original embeddings. The proposed method outperformed previous studies that adapt contextualized embeddings. It achieved state-of-the-art performance on knowledge-based WSD when combined with the reranking heuristic that uses the sense inventory. We found that the similarity characteristics of specialized embeddings conform to the key idea. We also found that the (dis)similarity of embeddings between the related/different/unrelated senses correlates well with the performance of WSD.

\end{abstract}

\section{Introduction}
\label{sec:introduction}

Word Sense Disambiguation (WSD) is the task of choosing the appropriate sense of a word from a given sense inventory using contextual information. WSD has proven its usefulness for 
Information Retrieval~\citep{DBLP:conf/acl/ZhongN12} and Machine Translation~\citep{DBLP:conf/acl/CampolungoMSN22}. A series of extensive studies has led supervised WSD task performance to surpass the milestone of 80\% accuracy~\citep{DBLP:conf/acl/BevilacquaN20}, which is the estimated human performance~\citep{DBLP:journals/csur/Navigli09}. 

In contrast, the goal of this study is \textit{knowledge-based WSD}: a variant of WSD that does not rely on supervision data but only on lexical knowledge (e.g., word ontology). This task setting is practically appealing because it does not use a corpus with sense annotations~\citep{DBLP:conf/ijcai/BevilacquaPRN21}, 
which is costly and labor-intensive to prepare.



A promising approach is based on similarity: to select the sense that is the nearest to a target word in the embedding space~\citep{DBLP:conf/emnlp/WangW20}. Specifically, a pre-trained language model, typically BERT~\citep{DBLP:conf/naacl/DevlinCLT19}, is used to compute \textit{sense embeddings} for definition sentences. Similarly, a target word is encoded into a \textit{context embedding} for a given sentence. Then, the model predicts the sense of the target word by finding the most similar sense embedding to the context.

The inherent challenge of the similarity-based approach is how we associate two different representations of word meanings, either by definition sentences or by words in context. Although the BERT embeddings capture the coarse-grained word meanings~\citep{DBLP:conf/nips/ReifYWVCPK19,DBLP:journals/coling/LoureiroRPC21}, there should be room for improvement. Notably, \citet{DBLP:conf/emnlp/WangW20} proposed $\mathtt{SREF}$, sense embedding adaptation by bringing semantically related senses closer. 
Extending their work, \citet{DBLP:conf/emnlp/WangZW21} proposed $\mathtt{COE}$, context embedding enhancement heuristics during inference using the document-level global contexts of the given sentence, and reported the best performance. Despite being effective, $\mathtt{COE}$ cannot be applied to stand-alone texts, e.g., short messages on social media 
or search queries, limiting its applicability. 

\begin{figure}[t]
\includegraphics[width=\columnwidth]{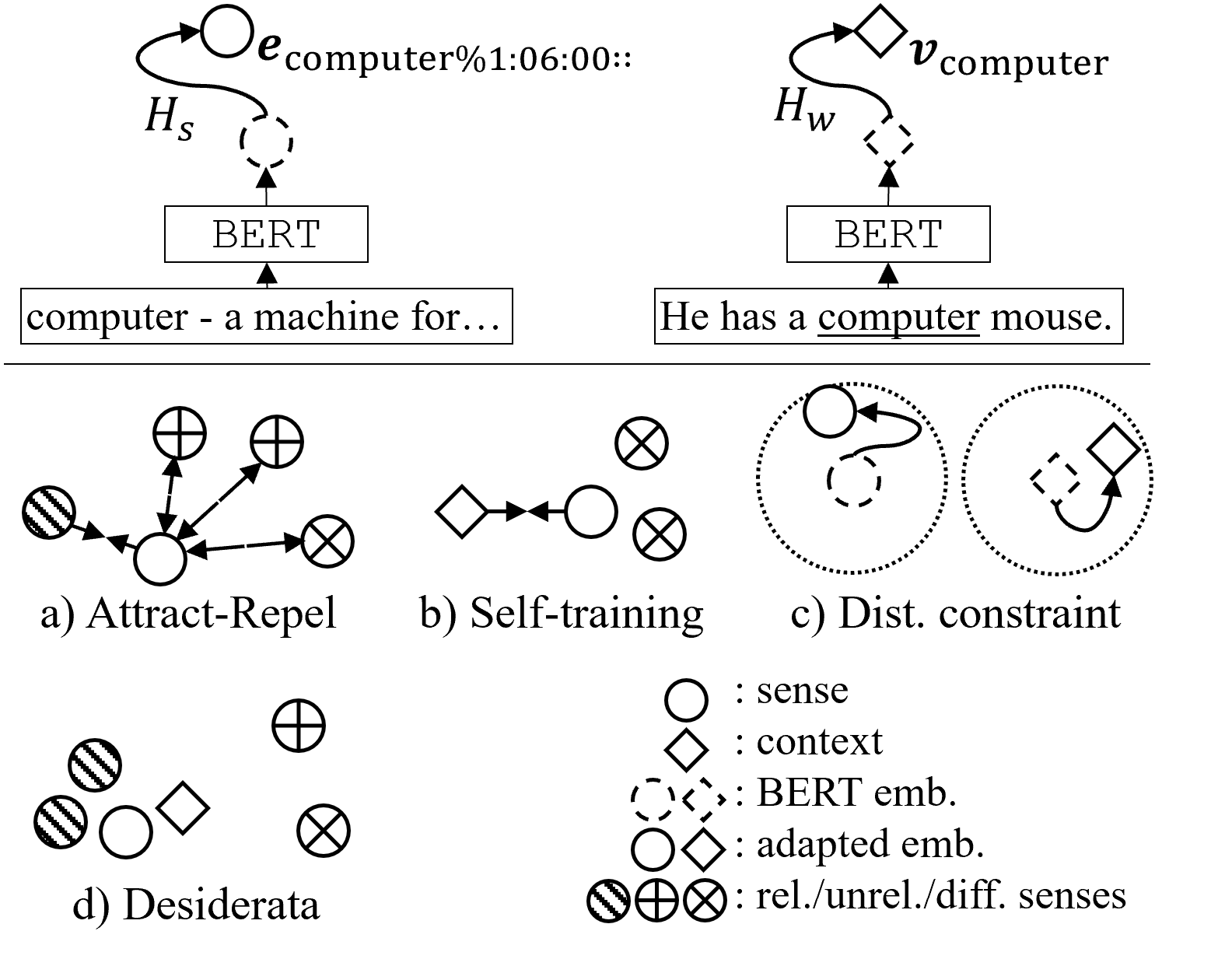}
\caption{\label{fig:schema_of_proposed_method}Schema of the proposed method. The BERT embeddings representing senses and contexts are adapted by transformation (top). Transformation functions are optimized using Attract-Repel and self-training objectives under distance constraints so that the adapted embeddings are effective for WSD (bottom).}
\end{figure}

Our study aims to improve both accuracy and applicability to stand-alone texts. 
Specifically, we propose an adaptation method of the sense and context embeddings for the WSD task solely using lexical knowledge. 
Then, what are good embeddings for WSD?
Our key idea is to 1) bring semantically related sense and context embeddings that convey the same meaning closer, and 2) send unrelated and/or different senses that share the same surface form farther away (Fig.~\ref{fig:schema_of_proposed_method}-d). We formulate the idea as the Attract-Repel objective and self-training objective. The main novelty is the joint optimization to utilize their complementary nature: the former should improve the distinguishability between senses whereas the latter offers pseudo signals of context-sense associations, which has not been explored in previous methods.

The Attract-Repel objective, inspired by~\citet{DBLP:conf/naacl/VulicM18}, injects semantic relation knowledge into the similarity of sense pairs.
Specifically, we make semantically related senses more similar while making different and unrelated senses more dissimilar (Fig.~\ref{fig:schema_of_proposed_method}-a). While $\mathtt{SREF}$ performs Attract only, our method utilizes both Attract and Repel. 

The self-training objective, inspired by the idea of retraining on the classifier's own predictions instead of annotated senses~\citep{DBLP:journals/csur/Navigli09}, updates the similarity of context-sense pairs in a pseudo labeling manner (\S~\ref{sec:preliminary-experiments}). Specifically, for each training step and given context, we bring the nearest neighbor sense among candidates closer (Fig.~\ref{fig:schema_of_proposed_method}-b). 
We also impose distance constraints during adaptation to control the deviation from BERT embeddings (Fig.~\ref{fig:schema_of_proposed_method}-c) because excessive deviation may cause an inaccurate nearest neighbor sense selection
, which would cause a performance drop.

We call the overall proposed method $\mathtt{SS}\text{-}\mathtt{WSD}$, Semantic Specialization for WSD, following \citet{DBLP:conf/naacl/VulicM18}. We evaluated $\mathtt{SS}\text{-}\mathtt{WSD}$ using the standard evaluation protocol~\citep{DBLP:conf/eacl/NavigliCR17} and confirmed that it outperforms the previous embeddings adaptation method. Furthermore, it achieved state-of-the-art (SoTA) performance when combined with the reranking heuristic that uses a sense inventory~\citep{DBLP:conf/acl/WangW20}, and thus is applicable to stand-alone texts. 

The contributions of our study are as follows:
\begin{itemize}
  \item We proposed $\mathtt{SS}\text{-}\mathtt{WSD}$, an embedding adaptation method that achieves new SoTA in knowledge-based WSD, regardless of the availability of document-level global contexts.
  \item We found that the performance gain originates from the joint optimization of Attract-Repel and self-training objectives and the prevention of deviation from the original embeddings.
  \item Empirically, we found that the similarity of related/different/unrelated senses \textit{relative to} the similarity of ground-truth context-sense pairs correlates well with the WSD performance.
\end{itemize}

\section{Related Work}

\subsection{Knowledge-based WSD}
Knowledge-based WSD is a variant of WSD that does not use a sense annotation corpora such as the SemCor~\citep{DBLP:conf/naacl/MillerLTB93} but uses lexical resources instead, typically WordNet. 
The majority vote based on sense frequencies, also known as the WordNet first sense heuristic~\citep{speechanddlanguage}, is a simple but strong baseline method of this category. Sense definitions and usage examples are also used to measure the similarity of the target word in a sentence. The simplest method is based on word overlap~\citep{DBLP:conf/sigdoc/Lesk86}.

One recent direction is the use of BERT as a contextualized encoder. BERT embeddings showed empirical success on the supervised WSD task when used as features. Some analyses reported that BERT embeddings capture the coarse-grained word meanings~\citep{DBLP:conf/nips/ReifYWVCPK19,DBLP:journals/coling/LoureiroRPC21}. 
\citet{DBLP:conf/emnlp/WangW20} proposed a similarity-based method in the embedding space. 
It chooses the sense which has the most similar embedding, formed from the concatenation of its lemma, definition, and usage examples, to the embedding of a target word.
They also proposed the Semantic Relation Enhancement Framework ($\mathtt{SREF_{emb}}$), which adapts sense embeddings by weighted averages over semantically related senses, e.g., hyponyms and derivations. $\mathtt{SREF_{emb}}$ is the most high-performing adaptation method so far. We report that our proposed method achieves better performance.

\subsection{Heuristics for Knowledge-based WSD}
Another recent direction is the heuristics for choosing the most similar sense, which is further divided into those that use the sense inventory information and those that exploit the document-level global contexts of a given sentence. \citet{DBLP:conf/emnlp/WangW20} proposed the former, the Try-again Mechanism (TaM). It  reranks candidates by adding the similarity between the target word and the lexicographer class (supersense) that a candidate sense belongs to. Subsequent studies~\citep{DBLP:conf/emnlp/WangZW21,DBLP:conf/acl/WangW20} refined TaM using Coarse Sense Inventory~\citep{DBLP:conf/aaai/LacerraBPN20}. We examine the effectiveness of the proposed method combined with TaM because it can be applied to stand-alone texts.

\citet{DBLP:conf/emnlp/WangZW21} proposed contextual information enhancement (CIE), which enhances context embeddings by exploiting the document-level global contexts of a given sentence on evaluation. This idea originally stems from the one-sense-per-discourse hypothesis~\citep{DBLP:conf/naacl/GaleCY92}: that the sense of a word is highly consistent within a document.

\subsection{Attract-Repel Framework}
The Attract-Repel Framework is used to inject lexical knowledge into embeddings by encouraging similar instances to have closer embeddings while encouraging dissimilar instances to be farther away. \citet{DBLP:conf/naacl/VulicM18} and \citet{DBLP:journals/tacl/MrksicVSLRGKY17} reported that updating static word embeddings using lexical knowledge improves the performance of the word-level semantic relation classification task.  Our study proposes its application to sense and context embeddings for the WSD task. We also reformulate the original loss function with the contrastive loss, inspired by its success in Computer Vision~\citep{DBLP:conf/icml/ChenK0H20} and NLP~\citep{DBLP:conf/emnlp/GaoYC21,DBLP:conf/acl/WangDLZ20,DBLP:conf/acl/GiorgiNWB20}.



\subsection{Supervised WSD}
\label{subsec:supervised_method}

Supervised methods rely on corpora of sense-annotated contexts, such as SemCor, for training models. However, the coverage of words and senses is limited and biased towards more frequent senses~\cite{DBLP:conf/ijcai/Pasini20}. Recent studies have addressed these limitations by incorporating lexical resources into the methods. 
\citet{DBLP:conf/naacl/BarbaPN21} and its subsequent study~\cite{DBLP:conf/emnlp/BarbaPN21} reframed WSD as a span extraction task by appending definition sentences of candidate senses to the target context. They reached the SoTA performance among supervised methods.

Similarity-based approaches are also used with supervised methods. Supervised k-nearest neighbors ($\mathtt{Sup}\text{-}\mathtt{kNN}$)~\citep{DBLP:conf/acl/LoureiroJ19} defines sense embeddings as the averaged context embeddings of annotated senses.
The Bi-Encoder model ($\mathtt{BEM}$)~\cite{DBLP:conf/acl/BlevinsZ20} jointly fine-tunes two BERT encoders for definition sentences and contexts, ensuring that context embeddings will be closer to the correct sense embeddings. The proposed method is similar in architectural design to $\mathtt{BEM}$, but differs in that we do not fine-tune the BERT encoders.
We will compare our results with $\mathtt{Sup}\text{-}\mathtt{kNN}$ and $\mathtt{BEM}$ to assess the effect of using no sense annotation and of freezing BERT encoders on performance.

\section{Semantic Specialization for WSD}
\label{sec:semantic_specialization_for_wsd}

\subsection{Formalization of WSD}
The proposed method adapts BERT embeddings by trainable transformation functions $H_s$ and $H_w$:
\begin{align}
    \mathbf{v}_w &= H_w(\mathbf{\hat v}_w), \\
    \mathbf{e}_s &= H_s(\mathbf{\hat e}_s),
\end{align}
where the inputs $\mathbf{\hat v}_w$ and $\mathbf{\hat e}_s$ are the context and sense embeddings computed by a BERT encoder and the outputs $\mathbf{v}_w$ and $\mathbf{e}_s$ are the specialized embeddings. 

We train the transformation functions by minimizing the weighted sum of the Attract-Repel objective and the self-training objective on the specialized embeddings. Note that the BERT encoder is frozen (not fine-tuned). We integrate the constraints on the distance between the input and output into the architecture of transformation functions (\S~\ref{subsec:transformation_function}).

To predict a sense for a given target word $w$, we look up the candidate senses $\mathcal{S}_w$ and compute their specialized sense embeddings using the learned function $H_s$. Similarly, we compute specialized context embeddings using $H_w$. Then, we select the nearest neighbor sense $s^{*}$ using cosine similarity:
\begin{align}
\label{eq:inference}
    s^{*} &= \argmax_{s' \in \mathcal{S}_w} \rho_{w,s'}, \\
    \rho_{w,s} &= \cos(\mathbf{v}_w, \mathbf{e}_s) = \frac{\mathbf{v}_w \cdot \mathbf{e}_s}{\|\mathbf{v}_w\| \|\mathbf{e}_s\|} .
\end{align}


\subsection{Lexical Knowledge in WordNet}
\label{subsec:lexical_resources}
We use WordNet~\cite{mitpress/Miller1998} as a lexical resource and sense inventory. WordNet mainly consists of synsets, lemmas, and senses. A synset is a group of synonymous words that convey a specific meaning.
A lemma presents a canonicalized form of a word and belongs to one or more synsets.
A sense is the lemma disambiguated by a sense key, and belongs to a single synset. We use the sense key as the identifier of a sense.

The proposed method makes use of relational knowledge between senses for training the transformation functions. Specifically, for each sense $s$, we collect three sets of senses: \textit{related} $\mathcal{S}^{P}_s$, \textit{different} $\mathcal{S}^{N}_s$, and \textit{unrelated} $\mathcal{S}^{U}_s$. 
The \textit{related} set consists of sense keys of synonyms and semantically related senses (e.g., hyponyms) to the target sense. We followed the definition of related senses used in \citet{DBLP:conf/emnlp/WangW20} (Appendix~\ref{appendix:lexical_resources}). 
The \textit{different} set consists of sense keys sharing the same lemma to the target sense excluding itself. In other words, the different senses correspond to the polysemy of the lemma of the target sense. 
The \textit{unrelated} set presents sense keys that are randomly chosen from the sense inventory (see \S~\ref{subsubsec:attract_repel_objective} for details).
Table~\ref{tbl:lexical_resources_stats} shows the statistics of lemmas and senses. See Table~\ref{tbl:lexical_resources} (in Appendix~\ref{appendix:lexical_resources}) for examples of the concepts explained in this subsection.


\begin{table}[t]
\setlength{\tabcolsep}{4pt}
\footnotesize
\centering
\begin{tabular}{lrrrrr}
\hline\hline
\multicolumn{1}{c}{Element} & \multicolumn{1}{c}{Noun} & \multicolumn{1}{c}{Verb} & \multicolumn{1}{c}{Adj.} & \multicolumn{1}{c}{Adv.} & \multicolumn{1}{c}{All} \\
\hline
\# Lemmas & 117,798 & 11,529  & 21,479  & 4,481   & 155,287 \\
\# Senses & 146,320 & 25,047  & 30,002  & 5,580   & 206,949 \\
Rel. senses & 7.8 & 13.0 & 6.2 & 3.9 & 8.1 \\
Diff. senses & 0.8 & 4.1 & 1.2 & 0.7 & 1.3 \\
\hline
\end{tabular}
\caption{\label{tbl:lexical_resources_stats}Summary statistics of lexical resources by part-of-speech tag. 
Values in the related and different senses rows indicate the average per sense.
}
\end{table}

\subsection{BERT Embeddings for Sense and Context}
\label{sec:contextualized-encoding}

For obtaining BERT embeddings, we follow the standard practice of the previous studies~\citep{DBLP:journals/kbs/WangWF20,DBLP:conf/acl/BevilacquaN20,DBLP:conf/emnlp/WangW20}. Specifically, we use \texttt{bert-large-cased}\footnote{We use \texttt{transformers} package~\cite{DBLP:conf/emnlp/WolfDSCDMCRLFDS20}.} with special tokens \texttt{[CLS]} and \texttt{[SEP]}. For each subword, we compute a sum over outputs at the last four layers of Transformer blocks.

A context embedding is the average of BERT embeddings over constituent subwords. For the computation of sense embeddings, we follow the method that \citet{DBLP:conf/emnlp/WangW20} used. 
See Appendix~\ref{appendix:bert_embeddings_for_sense} for details.

\subsection{Transformation Functions}
\label{subsec:transformation_function}
The proposed method adapts embeddings by applying the trainable transformation, i.e., the specialization is learned by optimizing the transformation functions. This approach enables the adaptation of context embeddings on the fly during inference, which was not possible in the original approach that directly learns adapted embeddings~\citep{DBLP:conf/naacl/VulicM18}.

Let $\mathbf{\hat{v}}_w$ and $\mathbf{\hat{e}}_s$ be context and sense BERT embeddings. We transform them independently using residual mapping functions $F_w$ and $F_s$, which are both two-layer feedforward networks, $\mathrm{FFNN}_w$ and $\mathrm{FFNN}_s$. These networks are comprised of a linear layer with a ReLU activation, followed by a linear layer with a sigmoid activation.
\begin{gather}
    \mathbf{v}_w = H_w(\mathbf{\hat v}_w) = \mathbf{\hat v}_w + \epsilon \|\mathbf{\hat v}_w\| F_w(\mathbf{\hat v}_w) \label{equ:h-w} ,\\
    \mathbf{e}_s = H_s(\mathbf{\hat e}_s) = \mathbf{\hat e}_s + \epsilon \|\mathbf{\hat e}_s\| F_s(\mathbf{\hat e}_s) \label{equ:h-s} ,\\
    F_w(\mathbf{\hat v}_w) = 2\sigma(\mathrm{FFNN}_w(\mathbf{\hat v}_w)) - 1, \\
    F_s(\mathbf{\hat e}_s) = 2\sigma(\mathrm{FFNN}_s(\mathbf{\hat e}_s)) - 1,
\end{gather}
where $\mathbf{v}_w$ and $\mathbf{e}_s$ are the specialized embeddings.
$\epsilon$ is the hyperparameter that controls how far away the specialized embeddings can be. Specifically, the L2 distance relative to the original embedding $\|\mathbf{v}_w - \mathbf{\hat v}_w\| / \|\mathbf{\hat v}_w\|$ 
is bounded by $\epsilon \sqrt {N_d}$, where ${N_d}$ is the dimension size of embeddings\footnote{${N_d}=1,024$ for \texttt{bert-large-cased}.}. This is because the residual functions map the inputs to the space $[-1, +1]^{N_d}$.


\subsection{Objectives}
\label{subsec:objective_functions}
We jointly optimize the Attract-Repel objective for sense pairs and the self-training objective for context-sense pairs by minimizing the weighted sum of the loss functions,
\begin{equation}
\label{eq:objective}
    L = L^{\mathrm{AR}} + \alpha L^{\mathrm{ST}},
\end{equation}
where $\alpha$ is the hyperparameter that determines the relative importance of the self-training objective.

The joint optimization is motivated by the complementary nature of these two objectives. The Attract-Repel objective should improve the separability of similar/different senses but does not contribute to determining which context and sense should be associated. In contrast, the self-training objective provides pseudo-supervision signals for context-sense associations, although the informativeness is, when used alone, limited because it essentially reinforces the similarity to the initial nearest neighbor sense of the target context (\S~\ref{subsubsec:self_training_objective}).

\subsubsection{Attract-Repel Objective}
\label{subsubsec:attract_repel_objective}
We formulate Attract-Repel objective loss $L^{\mathrm{AR}}$ using contrastive loss: we bring \textit{related} senses closer while \textit{different} and \textit{unrelated} senses farther away\footnote{In the contrastive learning literature, related, unrelated, and different senses correspond to the positives, weak negatives, and hard negative examples, respectively.} (\S~\ref{subsec:lexical_resources}). 
Specifically, for a given minibatch of senses $\mathcal{S}^{B}$ and a specific sense $s \in \mathcal{S}^{B}$, we define the subset excluding itself $\mathcal{S}^{B} \setminus \{s\}$ as the unrelated senses $\mathcal{S}^{U}_s$. Then, we randomly choose a sense $s_p$ from the related senses $\mathcal{S}^{P}_s$.
Similarly, we randomly choose up to five senses without replacement $\tilde{\mathcal{S}}^{N}_s$ from different senses $\mathcal{S}^{N}_s$.
Finally, $L^{\mathrm{AR}}$ for the minibatch $\mathcal{S}^B$ is defined as follows:

\begin{gather}
    L^{\mathrm{AR}} = - \sum_{s \in \mathcal{S}^{B}} \ln\frac{e^{\beta\rho_{s,s_p}}}{\sum\limits_{s' \in \left(\{s_p\} \cup \mathcal{S}^{U}_{s} \cup \tilde{\mathcal{S}}^{N}_{s}\right)}{e^{\beta\rho_{s,s'}}}} , \label{eq:objective_contrastive-loss} \\
    \rho_{s,s'} = \cos(\mathbf{e}_s, \mathbf{e}_{s'}).
\end{gather}
We set the scaling parameter $\beta$ to 64, following the suggestions in metric learning studies~\citep{DBLP:conf/cvpr/DengGXZ19,DBLP:conf/cvpr/WangWZJGZL018}. 


\subsubsection{Self-training Objective}
\label{subsubsec:self_training_objective}
We formulate the self-training objective loss $L^{\mathrm{ST}}$ so that we bring the contexts and nearest neighbor senses closer. In the self-training process, we label a word in context with the sense whose embedding is the closest to that of the word.
Specifically, let $\mathcal{W}^B$ denote a minibatch of words. For a word $w \in \mathcal{W}^B$, we obtain a set of candidate senses\footnote{Querying WordNet for a tuple of lemma and part-of-speech tag returns the candidate senses.} $\mathcal{S}_w$. Then, $L^{\mathrm{ST}}$ for the minibatch $\mathcal{W}^B$ is defined as,
\begin{gather}
\label{eq:objective_self-training}
    L^{\mathrm{ST}} = \sum_{w \in \mathcal{W}^B} (1 - \max_{s \in \mathcal{S}_{w}}\rho_{w,s}), \\
    \rho_{w,s} = \cos(\mathbf{v}_w, \mathbf{e}_s) .
\end{gather}
Note that the nearest neighbor sense for the same context changes during training as we update parameters of the transformation functions for embeddings. Our intention is to  bootstrap the performance, which was impossible in the ``static counterpart'', e.g., pseudo-labeling with the WordNet first sense heuristic. That is also a motivation of introducing the distance constraint in Eq. \ref{equ:h-w} and \ref{equ:h-s}: we were concerned about the performance drop when a large deviation occurs in the semantic specialization. We report empirical evidence that the constraint improves the performance (\S~\ref{subsec:ablation_epsilon}).


In principle, the training data can be any corpus annotated with lemmas and part-of-speech tags. Nevertheless, we used the SemCor~\citep{DBLP:conf/naacl/MillerLTB93} corpus with the sense annotations removed. This is because using these de-facto standard corpora contributes to better reproducibility and fairer comparisons.

\subsection{Try-again Mechanism (TaM) Heuristic}
\label{subsec:try-again-mechanism}
We examine the effectiveness of the proposed method when combined with TaM. Specifically, we employ the variant~\citep{DBLP:conf/acl/WangW20}\footnote{We followed author's implementation: \url{https://github.com/lwmlyy/SACE}} that utilizes Coarse Sense Inventory (CSI)~\citep{DBLP:conf/aaai/LacerraBPN20} because of its simplicity. In essence, TaM reranks candidate senses by updating similarities under the assumption that the context should be also similar to the coarse semantic category that the candidate sense belongs to. Let $s_1$ and $s_2$ be the top two nearest neighbors for the target word $w$ and $\mathcal{S}^{\mathrm{CSI}}_{s}$ be the set of senses\footnote{$\mathcal{S}^{\mathrm{CSI}}_{s}$ will be the empty set if $s$ doesn't exist in the CSI because it does not cover all synsets.} belonging to the same CSI class as $s$ belongs to. Then, we refine the similarity $\rho^{+}_{w,s}$ for each $s \in \{s_1,s_2\}$,
\begin{equation}
\label{eq:try-again-mechanism}
    \rho^{+}_{w,s} = \rho_{w,s} + \max_{s' \in \mathcal{S}^{\mathrm{CSI}}_{s}}{\rho_{w,s'}} .
\end{equation}
Finally, we choose the sense from $s_1$ and $s_2$ with highest similarity using $\rho^{+}_{w,s}$, i.e., we use the refined similarity $\rho^{+}_{w,s}$ instead of $\rho_{w,s}$ (Eq.~\ref{eq:inference}).

\section{Experiment Settings}

\subsection{Training}
We used WordNet senses for optimizing the Attract-Repel objective and the sense-annotated words in the SemCor corpus for the self-training objective. Note that we solely use lemmas and part-of-speech tags and disregard the sense annotations. The number of senses in WordNet is 206,949, and the number of words in the corpus is 226,036. We independently sampled minibatches $N_B$ for each objective. For the Attract-Repel objective, we iterate over all sense keys in the WordNet with 15 epochs\footnote{In each epoch, we discarded the remaining examples in the self-training objective trainset once all sense keys have been traversed.}.
For hyperparameter optimization, we disabled TaM heuristics and used the evaluation set of SemEval-2007 as the development set, following the standard practice~\citep{DBLP:conf/aaai/PasiniRN21}. See Appendix~\ref{appendix:hyperparameter_search} for details of the hyperparameter search. We set $N_B=256$, $\alpha=0.2$, and $\epsilon=0.015$. We used the Adam optimizer with learning rate $0.001$.

\subsection{Evaluation}
\label{subsec:evaluation}
For evaluation, we used the WSD unified evaluation framework~\citep{DBLP:conf/eacl/NavigliCR17}\footnote{Available at: \url{http://lcl.uniroma1.it/wsdeval/}}. We used the nearest neighbor sense as the prediction (Eq.~\ref{eq:inference}). For the evaluation metric, we adopt the micro-averaged F1 score\footnote{Note that F1 score is equal to Precision and Recall~\citep{DBLP:conf/aaai/PasiniRN21} because proposed method predicts a single sense.} that is commonly used in the literature. Unless otherwise specified, we run the training process five times with different random seeds, and report the mean and standard deviations.

\subsection{Baselines}
We compare the proposed method in two experimental configurations: \textit{Intrinsic} and \textit{With Heuristics}. For the \textit{Intrinsic} configuration, we compare it with the methods that do not use any heuristic. Specifically, we choose $\mathtt{PlainBERT}$ and $\mathtt{SREF_{emb}}$~\citep{DBLP:conf/emnlp/WangW20} as baselines. $\mathtt{PlainBERT}$ uses BERT embeddings $\mathbf{\hat v}_w$ and $\mathbf{\hat e}_s$ as is. 
$\mathtt{SREF_{emb}}$\footnote{We applied their method to $\mathtt{PlainBERT}$, consistent with the proposed method, to ensure a fair comparison of the effect of adaptation.} adapts sense embeddings so that it brings semantically related senses closer. For the \textit{With Heuristics} configuration, we compare the proposed method with the methods that combine heuristics. Specifically, we choose $\mathtt{SREF_{kb}}$~\citep{DBLP:conf/emnlp/WangW20} and $\mathtt{COE}$~\citep{DBLP:conf/emnlp/WangZW21} as baselines. $\mathtt{SREF_{kb}}$ combines $\mathtt{SREF_{emb}}$ with TaM. $\mathtt{COE}$ also utilizes $\mathtt{SREF_{emb}}$, but it employs refined TaM and CIE. 
$\mathtt{COE}$ is the current SoTA method on knowledge-based WSD.

We also compare with supervised methods which employ the similarity-based approach to assess the effect of not using sense annotations and of freezing BERT encoders. Specifically, we compare with $\mathtt{Sup}\text{-}\mathtt{kNN}$~\cite{DBLP:conf/acl/LoureiroJ19} and $\mathtt{BEM}$~\cite{DBLP:conf/acl/BlevinsZ20} (\S~\ref{subsec:supervised_method}), which both use SemCor as the trainset. $\mathtt{Sup}\text{-}\mathtt{kNN}$ computes sense embeddings as the context embeddings averaged over the annotated senses. $\mathtt{BEM}$ fine-tunes BERT encoders so that context embeddings and correct sense embeddings are brought closer.
We consider $\mathtt{BEM}$ as the de-facto upper bound of similarity-based approach, given its usage of a supervision signal to fine-tune the BERT encoders.

\section{Experimental Results}
\label{sec:experiment_results}

Table~\ref{tbl:main_results} shows the WSD task performance. In both configuration, the proposed method $\mathtt{SS}\text{-}\mathtt{WSD_{emb}}$ outperformed all knowledge-based baselines.

In the \textit{Intrinsic} configuration, $\mathtt{SS}\text{-}\mathtt{WSD_{emb}}$ outperformed $\mathtt{SREF_{emb}}$ by 3.9pt, which is as much as a 9.3pt improvement over $\mathtt{PlainBERT}$. Looking at the results for each part-of-speech, we observed the largest improvement over $\mathtt{SREF_{emb}}$ for verbs (9.0pt). This result reflects the fact that verbs have the richer supervision signal for the Attract-Repel objective because of the largest number of related and different senses (Table~\ref{tbl:lexical_resources_stats}) for verbs. This suggests that the richer semantic relation knowledge is, the higher performance the proposed method may achieve.

In the \textit{With Heuristics} configuration, $\mathtt{SS}\text{-}\mathtt{WSD_{kb}}$ outperformed $\mathtt{COE}$ by 0.8pt without using the CIE heuristic, which shows an advantage over the baselines regardless of whether the evaluation sentence is a stand-alone text or in a document.
The improvement brought by TaM was 2.2pt. Although $\mathtt{SS}\text{-}\mathtt{WSD_{kb}}$ lagged behind $\mathtt{COE}$ on the SE07 (SemEval-2007) subset, we think this result is understandable because $\mathtt{COE}$ also used SE07 for hyperparameter optimization. 

When compared to supervised methods, $\mathtt{SS}\text{-}\mathtt{WSD_{emb}}$ outperformed $\mathtt{Sup}\text{-}\mathtt{kNN}$ by 1.4pt, while falling behind $\mathtt{BEM}$ by 4.1pt. 
The results indicate that the proposed method associates contexts with senses more precisely than the example-based sense embeddings computation using sense-annotated contexts.
It also shows the effectiveness of the supervised fine-tuning of BERT encoders in $\mathtt{BEM}$, as evidenced through their ablation study~\cite{DBLP:conf/acl/BlevinsZ20}.

\begin{table*}[pt]
\setlength{\tabcolsep}{3pt}
\centering
\footnotesize
\begin{tabular}{lccrrrrrrrrrr}
\hline\hline
\multicolumn{1}{c|}{\multirow{2}{*}{ Method } } & 
\multicolumn{1}{c}{\multirow{2}{*}{ TaM } } & 
\multicolumn{1}{c|}{\multirow{2}{*}{ CIE } } & 
\multicolumn{5}{c|} {By subset} &
\multicolumn{4}{c|} {By part-of-speech} & 
\multicolumn{1}{c}{ \multirow{2}{*}{ All } } \\ \cline{4-12}

 \multicolumn{1}{c|}{} & \multicolumn{2}{c|}{} & \multicolumn{1}{c}{SE2}& \multicolumn{1}{c}{SE3}& \multicolumn{1}{c}{SE07} & \multicolumn{1}{c}{SE13} & \multicolumn{1}{c|}{SE15} & \multicolumn{1}{c}{Noun}& \multicolumn{1}{c}{Verb}& \multicolumn{1}{c}{Adj.} & \multicolumn{1}{c|}{Adv.} & \\ \hline


\multicolumn{13}{l}{Supervised}\\ \hline
\multicolumn{1}{l|}{ \makecell[l]{ $\mathtt{Sup}\text{-}\mathtt{kNN}$ \\ ~\citep{DBLP:conf/acl/LoureiroJ19} } } & × & \multicolumn{1}{c|}{×}  & 76.3& 73.2& 66.2& 71.7& \multicolumn{1}{r|}{74.1} & \multicolumn{1}{c}{---}  & \multicolumn{1}{c}{---}  & \multicolumn{1}{c}{---}  & \multicolumn{1}{c|}{---}  & 73.5 \\
\multicolumn{1}{l|}{ \makecell[l]{ $\mathtt{BEM}$ \\ ~\citep{DBLP:conf/acl/BlevinsZ20} } } & × & \multicolumn{1}{c|}{×} & 79.4& 77.4& \underline{74.5} & 79.7 & \multicolumn{1}{r|}{81.7} & 81.4 & 68.5 & 83.0 & \multicolumn{1}{r|}{87.9} & 79.0 \\ \hline
\multicolumn{13}{l}{Knowledge-based, \textit{Intrinsic} configuration}\\ \hline

\multicolumn{1}{l|}{$\mathtt{PlainBERT}$}& ×   & \multicolumn{1}{c|}{×}  & 67.8& 62.7& 54.5& 64.5& \multicolumn{1}{r|}{72.3} & 67.8& 52.3& 74.0& \multicolumn{1}{r|}{77.7} & 65.6\\
\multicolumn{1}{l|}{ \makecell[l]{$\mathtt{SREF_{emb}}$ \\ ~\citep{DBLP:conf/emnlp/WangW20}} } & ×   & \multicolumn{1}{c|}{×}  & 70.3& 68.0& 60.4& 74.2& \multicolumn{1}{r|}{77.4} & 76.3& 53.5& 75.2& \multicolumn{1}{r|}{76.3} & 71.0\\
\multicolumn{1}{l|}{\makecell[l]{$\mathtt{SS}\text{-}\mathtt{WSD_{emb}}$ (Ours)}} & ×   & \multicolumn{1}{c|}{×}  & \multicolumn{1}{c}{\makecell{\textbf{74.6}*\\ (0.5)}} & \multicolumn{1}{c}{\makecell{\textbf{73.0}*\\ (0.6)}} & \multicolumn{1}{c}{\makecell{\underline{\textbf{65.0}}*\\ (1.3)}} & \multicolumn{1}{c}{\makecell{\textbf{77.0}*\\ (0.5)}} & \multicolumn{1}{c|}{\makecell{\textbf{79.9}*\\ (1.0)}} & \multicolumn{1}{c}{\makecell{\textbf{78.2}*\\ (0.4)}} & \multicolumn{1}{c}{\makecell{\textbf{62.5}*\\ (0.7)}} & \multicolumn{1}{c}{\makecell{\textbf{79.7}*\\ (0.3)}} & \multicolumn{1}{c|}{\makecell{\textbf{80.5}*\\ (1.5)}} & \multicolumn{1}{c}{\makecell{\textbf{74.9}*\\ (0.3)}} \\ \hline
\multicolumn{13}{l}{Knowledge-based, \textit{With Heuristics} configuration} \\ \hline
\multicolumn{1}{l|}{ \makecell[l]{ $\mathtt{SREF_{kb}}$ \\ ~\citep{DBLP:conf/emnlp/WangW20} } } & \checkmark & \multicolumn{1}{c|}{×}  & 72.7& 71.5& 61.5& 76.4& \multicolumn{1}{r|}{79.5} & 78.5& 56.6& 79.0& \multicolumn{1}{r|}{76.9} & 73.5\\
\multicolumn{1}{l|}{ \makecell[l]{ $\mathtt{COE}$ \\ ~\citep{DBLP:conf/emnlp/WangZW21} } } & \checkmark & \multicolumn{1}{c|}{\checkmark} & 76.0& 74.2& \underline{\textbf{69.2}}& \textbf{78.2}& \multicolumn{1}{r|}{80.9} & \textbf{80.6}& 61.4& 80.5& \multicolumn{1}{r|}{81.8} & 76.3\\
\multicolumn{1}{l|}{$\mathtt{SS}\text{-}\mathtt{WSD_{kb}}$ (Ours)} & \checkmark  & \multicolumn{1}{c|}{×}  & \makecell{\textbf{77.7}*\\ (0.5)} & \makecell{\textbf{75.9}*\\ (0.6)} & \makecell{\underline{66.5} \\ (1.0)} & \makecell{78.0 \\ (0.5)} & \multicolumn{1}{r|}{\makecell{\textbf{81.6}\\ (0.9)}} & \makecell{79.3\\ (0.3)} & \makecell{\textbf{65.7}*\\ (0.8)} & \makecell{\textbf{84.9}*\\ (0.4)} & \multicolumn{1}{r|}{\makecell{\textbf{84.2}*\\ (0.8)}} & \makecell{\textbf{77.1}*\\ (0.3)} \\ \hline
\end{tabular}
\caption{\label{tbl:main_results}WSD performance by subset and part-of-speech tag. $\mathtt{SS}\text{-}\mathtt{WSD_{emb,kb}}$ are the proposed methods. Numbers in parentheses represent the standard deviation. Asterisks (*) indicate that the difference to the best baseline is statistically significant at $p<0.05$ by the Student's \textit{t}-test (two-tailed test). 
Checkmarks (\checkmark) in the TaM and CIE columns represent the usage of those heuristics. 
We bolded the best result among knowledge-based methods in each configuration and underlined the objective for hyperparameter tuning.
The scores of $\mathtt{BEM}$, $\mathtt{Sup}\text{-}\mathtt{kNN}$, $\mathtt{SREF_{kb}}$, and $\mathtt{COE}$ are taken from the original papers. 
}
\end{table*}

\section{Analysis}
\label{sec:analysis}

\subsection{Vanilla BERT Embeddings}
\label{sec:preliminary-experiments}
The proposed method adapts the BERT embeddings ($\mathtt{PlainBERT}$) by transformation. Therefore, its performance is influenced by the ability of $\mathtt{PlainBERT}$ to disambiguate senses. 

Table~\ref{tbl:plainbert_wsd} shows the WSD task performance using $\mathtt{PlainBERT}$. We also reported the WordNet first sense heuristic ($\mathtt{WN1^{st}Sense}$) for reference. We observe that $\mathtt{PlainBERT}$ is comparable to $\mathtt{WN1^{st}Sense}$, indicating that self-training is a more effective strategy than $\mathtt{WN1^{st}Sense}$ for obtaining pseudo sense labels.

\begin{figure}[t]
\includegraphics[width=\columnwidth]{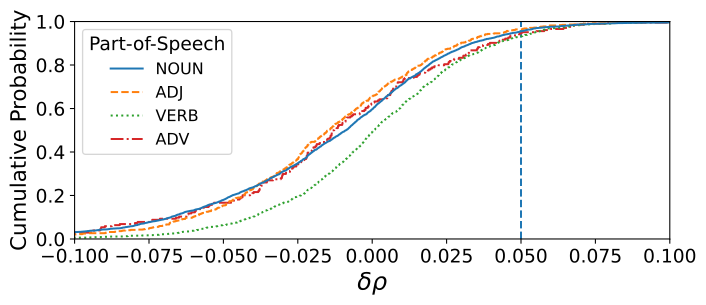}
\caption{\label{fig:similarity_gap}Cumulative distribution of the similarity margin between the incorrect sense and correct sense:
$\delta\rho_w = \max_{s' \in \mathcal{S}_w \backslash \mathcal{S}_w^{\mathrm{gt}}} \rho_{w,s'} - \max_{s' \in \mathcal{S}_w^{\mathrm{gt}}} \rho_{w,s'}$, where $\mathcal{S}^{\mathrm{gt}}_w$ is the set of ground-truth senses of the word $w$.}
\end{figure}

Fig.~\ref{fig:similarity_gap} shows the distribution of the similarity margin (difference) between the nearest neighbor incorrect sense and ground-truth sense computed by $\mathtt{PlainBERT}$. We used the evaluation set for this analysis. We found that the similarity margin is below 0.05 for approximately 90\% of all instances. This indicates that a large deviation from $\mathtt{PlainBERT}$ is not necessary for replacing nearest neighbor senses with the ground-truth ones. 

\begin{table}[t]
\centering
\footnotesize
\begin{tabular}{lc}
\hline\hline
\multicolumn{1}{c}{Method}    & WSD (All) \\ \hline
$\mathtt{WN1^{st}Sense}$ & 65.2 \\
$\mathtt{PlainBERT}$ & 65.6 \\ \hline 
\end{tabular}
\caption{\label{tbl:plainbert_wsd}F1 score of BERT embeddings ($\mathtt{PlainBERT}$) and WordNet the first sense heuristic ($\mathtt{WN1^{st}Sense}$).}
\end{table}

\subsection{Effect of Objectives}
\label{subsec:ablation_objectives}
Table~\ref{tbl:ablation_objective} shows the performance comparison when we eliminate a specific component from the semantic specialization objectives (\S~\ref{subsec:objective_functions}). We keep all hyperparameters unchanged. 

When we exclude either the Attract-Repel objective 
or the self-training objective, 
we see the performance drop by 3.3pt and 4.4pt, respectively. This finding supports the claim that joint optimization is crucial for its complementary nature. 

When we remove either the unrelated senses or different senses from the Attract-Repel objective, we also see the performance drop by 5.0pt and 1.4pt, respectively. This result supports the idea that bringing semantically unrelated and different senses farther away contributes to performance. 
We also find that unrelated senses are more effective than different senses. 
A possible cause is the number of examples: while the number of unrelated senses is always\footnote{Minibatch size (=256) minus one yields 255.} 255, the number of different senses is, on average, just 1.3 (see Table~\ref{tbl:lexical_resources_stats})\footnote{In fact, only 38\% of all senses have different senses.}.

Disabling the adaptation of context embeddings (by using identity transformation) caused a performance drop of 3.2pt, indicating that adapting both sense and context embeddings is necessary.

\begin{table}[t]
\centering
\footnotesize
\begin{tabular}{lrr}
\hline\hline
\multicolumn{1}{c}{Ablation} & \multicolumn{1}{c}{WSD (All)} & \multicolumn{1}{c}{$\Delta${[}pt{]}} \\
\hline
$\mathtt{SS}\text{-}\mathtt{WSD_{emb}}$ & 74.9 & \multicolumn{1}{c}{---} \\
\hline
-Attract-Repel \textit{objective} & 71.6 & -3.3 \\
-Self-training \textit{objective} & 70.5  & -4.4 \\
-Unrelated senses $\mathcal{S}^{U}$ \textit{repelling} & 69.9 & -5.0 \\
-Different senses $\mathcal{S}^{N}$ \textit{repelling} & 73.5 & -1.4 \\
-Context \textit{adaptation} & 71.7 & -3.2 \\
\hline
\end{tabular}
\caption{\label{tbl:ablation_objective}Ablation study of training objective. \textit{Objective} rows represent the corresponding objective is excluded. \textit{Repelling} rows represent the corresponding sense pairs are removed from the Attract-Repel objective (Eq.~\ref{eq:objective_contrastive-loss}). \textit{Adaptation} rows represent the usage of identity transformation.
All differences are statistically significant at $p<0.05$ by Welch's \textit{t}-test (two-tailed test).}
\end{table}

\begin{figure}[t]
\centering
\includegraphics[width=\linewidth]{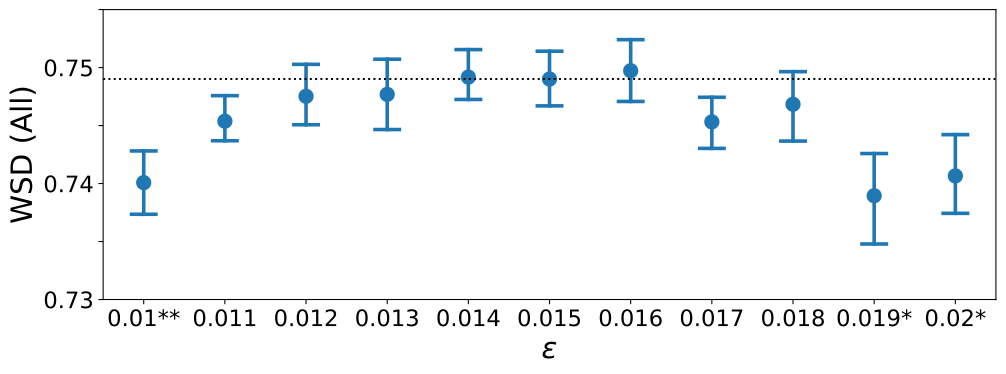}
\caption{\label{fig:ablation_epsilon_effect}
Ablation study of hyperparameter $\epsilon$
(\S~\ref{subsec:transformation_function}). 
Dot and error bar represent the mean and standard deviation, respectively. Horizontal line represents the default setting ($\epsilon=0.015$) performance. Asterisks indicate that the difference with respect to the default setting is statistically significant at $p<0.05$ (*) and $p<0.005$ (**) by Welch's \textit{t}-test (two-tailed test). 
}
\end{figure}

\subsection{Effect of Distance Constraint}
\label{subsec:ablation_epsilon}
Fig.~\ref{fig:ablation_epsilon_effect} shows the performance comparison when we change $\epsilon$, the hyperparameter that bounds how farther away the specialized embeddings can be, in the interval [0.01,0.02] with a step size of 0.001.
We found that performance follows an inverted U-shaped curve along $\epsilon$, indicating that a sweet spot exists. Briefly, it shows that a severe constraint (small $\epsilon$) results in an insufficient update for replacing nearest neighbors with ground-truth senses. In contrast, a looser constraint (large $\epsilon$) results in a substantial deviation, eventually making the self-training less effective in the training process. The latter fact supports the claim that controlling the deviation from the original embeddings is necessary.


\section{Effect of Self-training Dataset Size}
\label{appendix:analysis_self_training_dataset_size}

\begin{figure}[t]
\centering
\includegraphics[width=\linewidth]{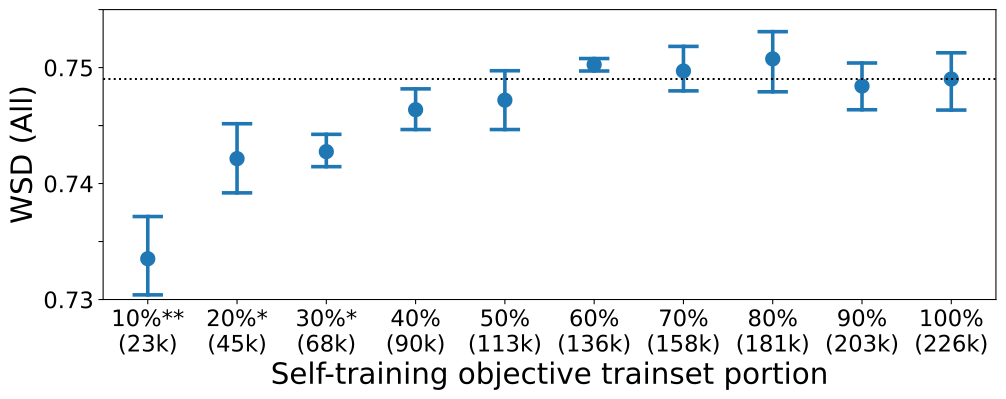}
\caption{\label{fig:ablation_self_training_portion}
Impact of varying the self-training dataset size from 10\% (23k examples) to 100\% (224k). The dot and error bar indicates the mean and standard deviation, respectively. The horizontal line represents the performance when utilizing the 100\% examples. Asterisks denote that the deviation from the 100\% is statistically significant at $p<0.05$ (*) and $p<0.005$ (**) by Welch's \textit{t}-test (two-tailed test).
}
\end{figure}

Fig.~\ref{fig:ablation_self_training_portion} illustrates the impact of varying the number of examples used for the self-training objective on the WSD task performance. It should be noted that 100\% in the figure corresponds to using all examples in the SemCor corpus. We found that performance improves as the number of examples increases and reaches a saturation point at 60\%, corresponding to 136k examples. While the coverage of words and senses appearing in the contexts also matters, it indicates that the benefits of self-training do not necessarily increase with the scaling to millions of examples.

\begin{table*}[ht]
\centering
\footnotesize
\begin{tabular}{l|rrrr|rrrr|r}
\hline\hline
\multicolumn{1}{c|}{Models} & \multicolumn{1}{c}{$\rho_{\mathcal{S}^{P}}$} & \multicolumn{1}{c}{$\rho_{\mathcal{S}^{U}}$} & \multicolumn{1}{c}{$\rho_{\mathcal{S}^{N}}$} & \multicolumn{1}{c|}{$\rho_{\mathcal{W}^{\mathrm{gt}}}$} & \multicolumn{1}{c}{$\Delta\rho_{\mathcal{S}^{P}}\uparrow$} & \multicolumn{1}{c}{$\Delta\rho_{\mathcal{S}^{U}}\downarrow$} & \multicolumn{1}{c}{$\Delta\rho_{\mathcal{S}^{N}}\downarrow$} & \multicolumn{1}{c|}{$\overline{\Delta\rho}\uparrow$} & \multicolumn{1}{c}{WSD (All)} \\ \hline
\multicolumn{1}{l|}{$\mathtt{PlainBERT}$} & 0.91 & 0.77   & 0.87   & 0.64  & 0.27 & 0.12   & 0.23   & -0.030 & 65.6   \\
\hline
\multicolumn{1}{l|}{$\mathtt{SS}\text{-}\mathtt{WSD_{emb}}$} & 0.88 & 0.64   & 0.78   & 0.77  & 0.11 & -0.13  & 0.01   & 0.078  & 74.9   \\
\multicolumn{1}{l|}{-Attract-Repel} & 0.92 & 0.79   & 0.90   & 0.81  & 0.11 & -0.02  & 0.08   & 0.014  & 71.6   \\
\multicolumn{1}{l|}{-Self-training}    & 0.88 & 0.64   & 0.78   & 0.61  & 0.27 & 0.02   & 0.17   & 0.027  & 70.5   \\
\multicolumn{1}{l|}{-Unrelated senses} & 0.90 & 0.73   & 0.79   & 0.73  & 0.17 & 0.00   & 0.06   & 0.033  & 69.9   \\
\multicolumn{1}{l|}{-Different senses} & 0.87 & 0.61   & 0.79   & 0.77  & 0.09 & -0.17  & 0.02   & 0.081  & 73.5   \\
-Context adaptation & 0.88 & 0.64   & 0.78   & 0.63  & 0.25 & 0.01   & 0.15   & 0.032  & 71.7  \\
\hline
\end{tabular}
\caption{\label{tbl:sense_context_similarity}Similarity characteristics of sense pairs and context-sense pairs. $\rho_{\mathcal{S}^{P}}$, $\rho_{\mathcal{S}^{U}}$, and $\rho_{\mathcal{S}^{N}}$ are the similarity to related, unrelated, and different senses (Eq.~\ref{eq:similarity_symbol_definition_senses}). $\rho_{\mathcal{W}^{\mathrm{gt}}}$ is the similarity of the context and its ground-truth senses (Eq.~\ref{eq:similarity_symbol_definition_contexts}). $\Delta\rho_{*}$ is the difference to $\rho_{\mathcal{W}^{\mathrm{gt}}}$ (Eq.~\ref{eq:similarity_relative_to_cs}). $\overline{\Delta\rho} = \frac{1}{3}(\Delta\rho_{\mathcal{S}^{P}}-\Delta\rho_{\mathcal{S}^{U}}-\Delta\rho_{\mathcal{S}^{N}})$. Uparrow$\uparrow$ (downarrow$\downarrow$) represents the positive (negative) direction is favorable. WSD (All) are replicated from Tables~\ref{tbl:main_results} and~\ref{tbl:ablation_objective} for reference.}
\end{table*}

\subsection{Similarity Characteristics}

We quantitatively investigate how well the proposed method achieved the key idea (Fig.~\ref{fig:schema_of_proposed_method}-d): bringing related senses and contexts closer while unrelated and different senses farther away. Specifically, in Table ~\ref{tbl:sense_context_similarity}, we reported averages of similarity values between related senses $\rho_{\mathcal{S}^{P}}$, unrelated senses $\rho_{\mathcal{S}^{U}}$, and different senses $\rho_{\mathcal{S}^{N}}$, along with averages of similarity values between ground-truth context-sense pairs\footnote{We used sense-annotated words in the evaluation dataset.} $\rho_{\mathcal{W}^{\mathrm{gt}}}$. See Appendix~\ref{appendix:analysis_similarity_characteristics} for formal definitions. We found that the proposed method $\mathtt{SS}\text{-}\mathtt{WSD_{emb}}$ brought context-sense pairs closer than $\mathtt{PlainBERT}$ ($\rho_{\mathcal{W}^{\mathrm{gt}}}$: $0.64 \rightarrow 0.77$). In contrast, it pushed the unrelated and different senses away: $\rho_{\mathcal{S}^{U}}$:$0.77 \rightarrow 0.64$ and $\rho_{\mathcal{S}^{N}}$:$0.87 \rightarrow 0.78$. These results demonstrate that joint optimization of the Attract-Repel and self-training objectives realized the key idea successfully.

Can we expect better performance if we realize the key idea more precisely? We investigated the relationship between these similarity metrics and WSD task performance. Specifically, we subtract $\rho_{\mathcal{W}^{\mathrm{gt}}}$ from each metric in order to capture the closeness of senses \textit{relative to} the correct context-sense pairs, defining $\Delta\rho_{*}$ as $\rho_{*} - \rho_{\mathcal{W}^{\mathrm{gt}}}$. For example, $\Delta\rho_{\mathcal{S}^{N}} = \rho_{\mathcal{S}^{N}} - \rho_{\mathcal{W}^{\mathrm{gt}}}$ should be a negative value because the average similarity among different senses $\rho_{\mathcal{S}^{N}}$ should be smaller than that among correct context-sense pairs $\rho_{\mathcal{W}^{\mathrm{gt}}}$. Therefore, we compute the value $\overline{\Delta\rho} = \frac{1}{3}(\Delta\rho_{\mathcal{S}^{P}}-\Delta\rho_{\mathcal{S}^{U}}-\Delta\rho_{\mathcal{S}^{N}})$ to estimate the WSD performance.

\begin{figure}[t]
\centering
\includegraphics[width=7.5cm]{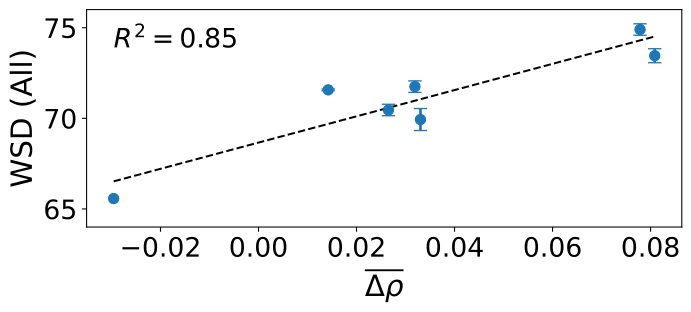}
\caption{\label{fig:similarty_metric_vs_wsd_performance}The relationship between the similarity characteristic metric $\overline{\Delta\rho}$ and WSD performance in Table~\ref{tbl:sense_context_similarity}.}
\end{figure}

Fig.~\ref{fig:similarty_metric_vs_wsd_performance} shows that $\overline{\Delta\rho}$ correlates well with WSD task performance ($R^2=0.85$). It suggests that if we achieve the key idea more precisely, we may improve the WSD performance.
For instance, using a richer lexical relation knowledge, exploitation of the monosemous words, and self-training with confidence thresholding may be promising. We leave it for future work.

\section{Conclusion}
In this paper, we proposed $\mathtt{SS}\text{-}\mathtt{WSD}$: Semantic Specialization for WSD\footnote{The source code is available at: \url{https://github.com/s-mizuki-nlp/semantic_specialization_for_wsd}}. The proposed method learns how to adapt BERT embeddings by transformation and uses the semantic relation knowledge 
as a supervision signal. The key idea is the desired characteristics of similarities: bringing related senses and the contexts 
closer while unrelated senses and different senses 
farther away. We realized it as the joint optimization of the Attract-Repel and self-training objectives while preventing large deviations from original embeddings.
Experiments showed that the proposed method outperformed the previous embedding adaptation method. When combined with the reranking heuristic that can be applied to stand-alone texts, it established a new SoTA performance on knowledge-based WSD. The proposed method performs well regardless of the availability of global contexts beyond the target sentence during inference, which the previous study did not achieve.
Several analyses showed the effectiveness of the objectives and constraints introduced for specialization. We also found that the closeness of semantically related/different/unrelated senses relative to the closeness of correct context-sense pairs positively correlates with the WSD task performance. 

\section{Future Work}
\label{sec:future_work}

Given that the proposed method only necessitates lexical resources, it has the potential to effectively address the knowledge acquisition bottleneck problem~\cite{DBLP:conf/ijcai/Pasini20}. Thus, we are interested in applying the proposed method to multilingual WSD using multilingual language models as contextualized encoders. One approach is the zero-shot cross-lingual transfer, which involves learning embeddings adaptation using only English lexical resources. Another option is the joint training of all target languages using multilingual lexical resources such as BabelNet~\cite{DBLP:conf/ijcai/NavigliBCMC21}.
We are also interested in integrating the proposed method into supervised WSD and applying the transfer learning of the specialized embeddings to other NLP tasks.

\section{Limitations}
\label{sec:limitations}
One limitation of this work is that it is specific to BERT. Although this is in line with the standard practice in previous studies, experimenting with other pre-trained language models is preferred to assess the utility of the proposed method, or to improve the performance further. Another limitation is that it is evaluated on a single dataset and task. While we also followed the de-facto standard protocol, evaluating on rare senses~\citep{DBLP:conf/acl/MaruCBN22} or Word-in-Context task~\citep{DBLP:conf/naacl/PilehvarC19,DBLP:conf/semeval/MartelliKTN21} will bring us more comprehensive insights on the effectiveness and applicability.

\section{Ethics Statement}
\label{sec:ethics_statement}
This work does not involve the presentation of a new dataset, nor the utilization of demographic or identity characteristics in formation.
In this work, we propose a method for adapting contextualized embeddings for WSD using lexical resources.
The proposed method is not limited to a specific resource, we used WordNet as the source of semantic relation knowledge and sense inventory.
Therefore, adapted embeddings and sense disambiguation behavior may reflect the incomplete lexical diversity of WordNet in culture, language~\citep{DBLP:conf/emnlp/0001BPRCE21}, and gender~\citep{DBLP:conf/wordnet/HicksRFB16}.

\section{Acknowledgments}
This work was supported by JSPS KAKENHI Grant Number 19H01118. We thank Marco Cognetta for his valuable input and for reviewing the manuscript.

\bibliography{custom}
\bibliographystyle{acl_natbib}

\clearpage

\appendix

\section{Lexical Resources}
\label{appendix:lexical_resources}
Table~\ref{tbl:lexical_resources} shows an example of lexical resources for a sense key \texttt{computer\%1:06:00::}. Note that unrelated senses are randomly chosen in practice.

For \textit{related} senses lookup, we followed \citet{DBLP:conf/emnlp/WangW20}'s paper and implementation\footnote{\url{https://github.com/lwmlyy/SREF}}. Briefly, for a given sense key, we collect the synsets that encompass either itself or the sense keys connected by \texttt{derivationally\_related\_forms} relation. Then, for each collected synset, we extend the synsets via semantic relations shown in Table~\ref{tbl:semantic_relations}. Finally, we collect the sense keys that belong to either one of the synsets in the extended set of synsets, together with those connected to a given sense key by semantic relations shown in Table~\ref{tbl:semantic_relations}.
We used the \texttt{nltk.corpus.wordnet} package for implementation.

\section{BERT Embeddings for Sense}
\label{appendix:bert_embeddings_for_sense}
For the computation of sense embeddings, we followed \citet{DBLP:conf/emnlp/WangW20}'s method.
Specifically, for a given sense key, we generate a sentence by filling in the following template using the lemma, synset lemmas, definition, and examples:
\begin{verbatim}
[lemma] - [syn. lemma 1], ...,
[syn. lemma n] - [definition]
[example 1] ... [example m],
\end{verbatim}
where \texttt{n} and \texttt{m} represent the number of synonym lemmas and the number of examples.
Then we take the average over all subwords in a sentence. For example, applying the template to the sense \texttt{computer\%1:06:00::} will produce the following sentence.

\noindent\fbox{%
    \parbox{\linewidth}{%
        computer - computer, computing device, data processor, ... - a machine for performing calculations automatically
    }%
}

We solely use the examples available in WordNet Gloss Corpus and do not use the augmented examples that \citet{DBLP:conf/emnlp/WangW20} collected.

\section{Hyperparameter Search}
\label{appendix:hyperparameter_search}
For the hyperparameter search, we first jointly optimized on the number of minibatches $N_B$, relative importance between objectives $\alpha$ (Eq.~\ref{eq:objective}), and constraint on the distance from BERT embeddings $\epsilon$ (Eq.~\ref{subsec:objective_functions}). We used \texttt{TPESampler} in the \texttt{optuna} package~\citep{DBLP:conf/kdd/AkibaSYOK19} for optimization. 
We run hyperparameter search over $N_B \in \{64,128,256,512,1024\}$, $\alpha \in [0.1, 10]$, and $\epsilon \in [0.001, 0.1]$. The number of search trials is 210.
Then, we ran a grid search on $\epsilon$ over the interval in [0.01,0.02] using a step size of 0.001. During hyperparameter search, we observed that 1) large minibatch size of 256 or above doesn't produce any statistically significant difference and 2) $\alpha$ is much less sensitive compared to $\epsilon$.

\begin{table*}[ht]
\centering
\begin{tabular}{lll}
\hline\hline
\multicolumn{1}{c}{\textbf{Element}} & \multicolumn{1}{c}{\textbf{Example}} \\
\hline
Sense (sense key) & \texttt{computer\%1:06:00::}\\
Lemma & \emph{computer} \\
Synset & \texttt{computer.n.01} \\
Definition sentence & \emph{a machine for performing calculations automatically} \\
Example & \texttt{Not Available} \\
Synonym lemmas & \emph{computer, computing device, data processor, ...} \\
Related senses & \makecell[l]{\texttt{computing\_device\%1:06:00::} (synonym),\\\texttt{analog\_computer\%1:06:00::} (hyponym),\\ \texttt{compute\%2:31:00::} (derivative), ...} \\
Different senses & \texttt{computer\%1:18:00::}\\
\makecell[l]{unrelated senses\\(randomly chosen)} & \makecell[l]{\texttt{goldfish\%1:05:00::}, \texttt{chef\%1:18:01::}, ...}\\
\hline
\end{tabular}
\caption{\label{tbl:lexical_resources}
Example of WordNet lexical resources used in the proposed method.}
\end{table*}

\begin{table*}[t]
\centering
\small
\begin{tabular}{ll}
\hline\hline
\makecell[c]{Category} & \makecell[c]{Relation names} \\
\hline
Sense key & \texttt{pertainyms, antonyms} \\
\hline
Synset & \makecell[l]{\texttt{hyponyms, hypernyms, part\_holonyms, part\_meronyms, member\_holonyms,} \\ \texttt{member\_meronyms, entailments, attributes, similar\_tos, causes,} \\ \texttt{substance\_holonyms, substance\_meronyms, usage\_domains, also\_sees}} \\
\hline
\end{tabular}
\caption{\label{tbl:semantic_relations}
WordNet semantic relation names used for collecting related senses.}
\end{table*}

\section{Analysis of Similarity Characteristics}
\label{appendix:analysis_similarity_characteristics}
We quantify the similarity characteristic as the macro average of similarity between senses and the similarity of ground-truth context-sense pairs. Specifically, for a given sense $s$, we calculate the average similarity to its related senses $\mathcal{S}^{P}_s$, unrelated senses $\mathcal{S}^{U}_s$, and different senses $\mathcal{S}^{N}_s$. Following Attract-Repel objective (\S~\ref{subsubsec:attract_repel_objective}), we define the minibatch excluding itself as the unrelated senses: $\mathcal{S}^{U}_s = \mathcal{S}^{B} \setminus \{s\}$. Then, we take the average over all senses $\mathcal{S}$, yielding the similarity among related senses $\rho_{\mathcal{S}^{P}}$, unrelated senses $\rho_{\mathcal{S}^{U}}$, and different senses $\rho_{\mathcal{S}^{N}}$ as follows:

\begin{equation}
\begin{aligned}
\label{eq:similarity_symbol_definition_senses}
    \rho_{\mathcal{S}^{P}} &= \frac{1}{|\mathcal{S}|}\sum_{s \in \mathcal{S}}\frac{1}{|\mathcal{S}^{P}_s|}\sum_{s' \in \mathcal{S}^{P}_s}{\rho_{s,s'}}, \\
    \rho_{\mathcal{S}^{U}} &= \frac{1}{|\mathcal{S}|}\sum_{s \in \mathcal{S}}\frac{1}{|\mathcal{S}^{U}_s|}\sum_{s' \in \mathcal{S}^{U}_s}{\rho_{s,s'}}, \\
    \rho_{\mathcal{S}^{N}} &= \frac{1}{|\mathcal{S}^{N}|}\sum_{s \in \mathcal{S}^{N}}\frac{1}{|\mathcal{S}^{N}_s|}\sum_{s' \in \mathcal{S}^{N}_s}{\rho_{s,s'}},
\end{aligned}
\end{equation}
where $\mathcal{S}^{N} = \{s;|\mathcal{S}^N_s|>0\}$.

For the similarity of ground-truth context-sense pairs $\rho_{\mathcal{W}^{\mathrm{gt}}}$, we use the pairs of the word and annotated senses in the evaluation dataset (\S~\ref{subsec:evaluation}). For a given word $w$, we calculate the average similarity to its ground-truth senses $\mathcal{S}^{\mathrm{gt}}_w$. Then, we take the average over all words $\mathcal{W}$ as follows:

\begin{equation}
\label{eq:similarity_symbol_definition_contexts}
    \rho_{\mathcal{W}^{\mathrm{gt}}} = \frac{1}{|\mathcal{W}|}\sum_{w \in \mathcal{W}}\frac{1}{|\mathcal{S}^{\mathrm{gt}}_w|}\sum_{s \in \mathcal{S}^{\mathrm{gt}}_w}{\rho_{w,s}}.
\end{equation}

Finally, we define $\Delta\rho_{*}$ as the difference to $\rho_{\mathcal{W}^{\mathrm{gt}}}$ for each relation types. We also define $\overline{\Delta\rho}$ as the arithmetic average over them while taking favorable positive/negative directions into account.

\begin{equation}
\begin{aligned}
\label{eq:similarity_relative_to_cs}
    \Delta\rho_{\mathcal{S}^{P}} &= \rho_{\mathcal{S}^{P}} - \rho_{\mathcal{W}^{\mathrm{gt}}} \\
    \Delta\rho_{\mathcal{S}^{U}} &= \rho_{\mathcal{S}^{U}} - \rho_{\mathcal{W}^{\mathrm{gt}}} \\
    \Delta\rho_{\mathcal{S}^{N}} &= \rho_{\mathcal{S}^{N}} - \rho_{\mathcal{W}^{\mathrm{gt}}} \\
    \overline{\Delta\rho} &= \frac{1}{3}(\Delta\rho_{\mathcal{S}^{P}}-\Delta\rho_{\mathcal{S}^{U}}-\Delta\rho_{\mathcal{S}^{N}})
\end{aligned}
\end{equation}

\section{Implementation Details}
We implemented the transformation functions using PyTorch library\footnote{\url{https://pytorch.org/}}. We trained them using single NVIDIA 2080Ti GPU. It took approximately two hours for a single run. We precomputed BERT embeddings for training and evaluation dataset and saved them to temporary files for computation efficiency.

\end{document}